%% file: main.tex
\documentclass[10pt,twocolumn,letterpaper]{article}

\usepackage{cvpr}

\usepackage{graphicx}
\usepackage{amsmath}
\usepackage{amssymb}
\usepackage{booktabs}

\usepackage{times}
\usepackage{epsfig}
\usepackage{graphicx}
\usepackage{amsmath}
\usepackage{amssymb}

\usepackage{subcaption} 
\usepackage{booktabs}    
\usepackage[font=small]{caption}
\usepackage{multirow}
\usepackage{multicol}
\usepackage{arydshln}
\usepackage{capt-of}
\usepackage{algorithm}
\usepackage{listings}
\usepackage{tabularx}
\usepackage{enumitem}
\usepackage[table]{xcolor}

\usepackage[accsupp]{axessibility}  

\definecolor{newlightblue}{RGB}{0,75,255}
\usepackage[pagebackref,breaklinks,colorlinks, citecolor={newlightblue}]{hyperref}

\usepackage[capitalize]{cleveref}
\crefname{section}{Sec.}{Secs.}
\Crefname{section}{Section}{Sections}
\Crefname{table}{Table}{Tables}
\crefname{table}{Tab.}{Tabs.}

\input{sections/commands}

\def\cvprPaperID{714}
\def\confName{CVPR}
\def\confYear{2022}

\begin{document}
\definecolor{lightblue}{RGB}{0,75,255}

\title{Comparing Correspondences: Video Prediction with Correspondence-wise Losses}

\author{
Daniel Geng \qquad
Max Hamilton \qquad
Andrew Owens \vspace{.8em}\\
University of Michigan~~~~\\
}

\maketitle

\input{sections/0_abstract}

\input{sections/1_intro_v3}

\input{sections/2_related_work}

\input{sections/3_method}

\input{sections/4_experiments}

\input{sections/5_discussion}

{\small
\bibliographystyle{ieee_fullname.bst}
\bibliography{main.bib}
}

\clearpage
\input{sections/supplementary}

\end{document}

%% file: sections/commands.tex
\DeclareMathOperator*{\argmin}{argmin}
\DeclareMathOperator*{\Ex}{\mathbb{E}}

\newcommand{\fig}[1]{Figure~\ref{#1}}

\newcommand{\tbl}[1]{Table~\ref{#1}}
\newcommand{\eqn}[1]{Equation~\ref{#1}}
\newcommand{\sect}[1]{Section~\ref{#1}}
\newcommand{\mypar}[1]{\vspace{-4mm}\paragraph{#1}}

\newcommand{\im}{{\mathbf{x}}}
\newcommand{\imgt}{\im}
\newcommand{\imest}{{\hat {\im}}}
\newcommand{\flow}{{\mathbf{u}}}
\newcommand{\flowval}{F}
\newcommand{\flowfn}{\mathcal{F}}

\newcommand{\loss}{\mathcal{L}}

\definecolor{lightyellow}{RGB}{255,255,170}

\newcommand{\supparxivcr}[3]{#2}

\usepackage{etoolbox}
\makeatletter
\AfterEndEnvironment{algorithm}{\let\@algcomment\relax}
\AtEndEnvironment{algorithm}{\kern2pt\hrule\relax\vskip3pt\@algcomment}
\let\@algcomment\relax
\newcommand\algcomment[1]{\def\@algcomment{\footnotesize#1}}
\renewcommand\fs@ruled{\def\@fs@cfont{\bfseries}\let\@fs@capt\floatc@ruled
  \def\@fs@pre{\hrule height.8pt depth0pt \kern2pt}%
  \def\@fs@post{}%
  \def\@fs@mid{\kern2pt\hrule\kern2pt}%
  \let\@fs@iftopcapt\iftrue}
\makeatother

%% file: sections/0_abstract.tex
\begin{abstract}


Image prediction methods often struggle on tasks that require changing the positions of objects, such as video prediction, producing blurry images that average over the many positions that objects might occupy.  In this paper, we propose a simple change to existing image similarity metrics that makes them more robust to positional errors: we match the images using optical flow, then measure the visual similarity of corresponding pixels. This change leads to crisper and more perceptually accurate predictions, and does not require modifications to the image prediction network. We apply our method to a variety of video prediction tasks, where it obtains strong performance with simple network architectures, and to the closely related task of video interpolation. Code and results are available at our webpage: \url{https://dangeng.github.io/CorrWiseLosses}

\end{abstract}

\input{tables/teaser}

%% file: tables/teaser.tex
\vspace{-10pt}
\begin{figure}
    \centering
    \includegraphics[width=\linewidth]{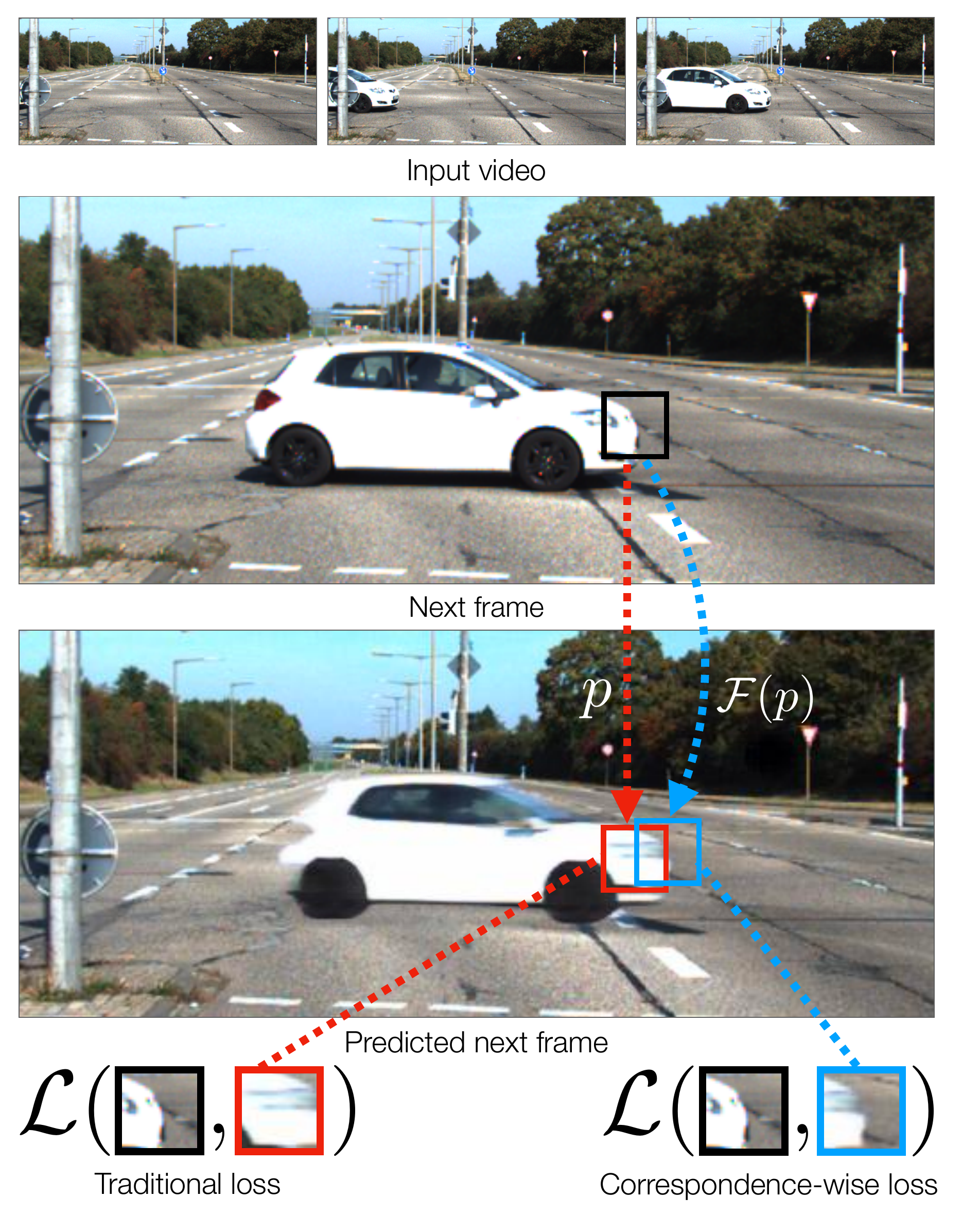}
    \vspace{-8mm}
    \caption{{\bf Correspondence-wise losses.} We propose a similarity metric that provides robustness
      to small positional errors, and apply it to image generation. We
      put the predicted and ground truth images into correspondence
      via optical flow, then measure the similarity between each pixel
      $p$ and its matching pixel $\mathcal{F}(p)$. Our metric leads to
      crisp predictions; it penalizes blurry, hedged images, like the
      one shown here, since they cannot be easily put into
      correspondence with the ground truth.}
    \label{fig:teaser}
    \vspace{-18pt}
\end{figure}

%% file: sections/1_intro_v3.tex
\section{Introduction}

Recent years have seen major advances in image prediction~\cite{isola2016pix2pix, chen2017CRN, park2019spade, karras2018stylegan, brock2018biggan}, yet these methods often struggle to successfully alter {\em image structure}. Consequently, tasks that involve modifying the positions or shapes of objects, such as video prediction and the closely related problem of video interpolation, remain challenging open problems. 

Often, there is fundamental uncertainty over where exactly an object should be. When this happens, models tend to produce blurry results. This undesirable behavior is often encouraged by the loss function. Under simple {\em pixel-wise} loss functions, such as the $L_1$ distance, each incorrectly positioned pixel is compared to a pixel that belongs to a different object, thereby incurring a large penalty. Models trained using these losses therefore ``hedge'' by averaging over all of the possible positions an object might occupy, resulting in images with significantly lower loss.

We take inspiration from classic image matching methods, such as Hausdorff matching~\cite{barrow1977parametric,huttenlocher1993comparing} and deformable parts models~\cite{fischler1973representation,felzenszwalb2005pictorial,felzenszwalb2009object}, that address this problem by allowing input images to undergo small spatial deformations before comparison. Before measuring the similarity of an image and a template, these methods first geometrically align them, thereby obtaining robustness to small variations in position or shape. 

Analogously, we propose a simple change to existing losses that makes them more robust to small positional errors. When comparing two images, we put them into correspondence using optical flow, then measure the similarity between {\em matching pairs of pixels}. Comparisons between the images therefore occur between pixel correspondences, rather than pixels that reside in the same spatial positions (\fig{fig:teaser}), \ie the loss is computed {\it correspondence-wise} rather than pixel-wise.

Despite its simplicity, our proposed ``loss extension" leads to crisper and more perceptually accurate  predictions. To obtain low loss, the predicted images must be easy to match with the ground truth via optical flow:
every pixel in a target image requires a high-quality match in the predicted image, and vice versa.
Blurry predictions tend to obtain high loss, since there is no simple, smooth flow field that puts them into correspondence with the ground truth. The loss also encourages objects to be placed in their correct positions, as positional mistakes lead to poor quality matches and occluded content, both of which incur penalties.

Since optical flow matching occurs within the loss function our approach does not require altering the design of the network itself. This is in contrast to popular {\em flow-based} video prediction architectures~\cite{zhou2016flow,liu2017video,gao2019disentangling,jaderberg2015spatial} that produce a deformation field within the network, then generate an image by  warping the input frames. 

We demonstrate the effectiveness of our method in a number of ways:
\begin{itemize}[leftmargin=5mm,topsep=0pt, noitemsep]
\item We show through experiments on a variety of video prediction tasks that our method significantly improves perceptual image quality. Our evaluation studies a variety of loss functions, including $L_1$ and $L_2$ distance and perceptual losses~\cite{gatys2016image,johnson2016perceptual}. These losses produce better results when paired with correspondence-wise prediction on egocentric driving datasets~\cite{geiger2013IJRR,Cordts2016Cityscapes}. 
\item We obtain video prediction results that outperform a flow-based state-of-the-art video prediction method~\cite{wu2020future} on perceptual quality metrics for KITTI~\cite{geiger2013IJRR} and Cityscapes~\cite{Cordts2016Cityscapes}, despite using a simple, off-the-shelf network architecture.
\item We apply our loss to the closely-related task of video interpolation~\cite{kalluri2021flavr}, where we obtain significantly better results than an $L_1$ loss alone. 
\item We show that our method also improves performance of stochastic, variational autoencoder (VAE) video prediction architectures~\cite{denton2018stochastic,villegas2019high}.

\end{itemize}

%% file: sections/2_related_work.tex
\section{Related Work}

\paragraph{Video prediction.} 

Early work in video prediction used recurrent networks to model long-range dependencies~\cite{sutskever2009recurrent, mittelman2014structured, michalski2014modeling,srivastava2015unsupervised, kalchbrenner2017video}. More recent work has focused on photorealistic video prediction using large convolutional networks. Lotter~\etal~\cite{lotter2016predictive} proposed a  predictive coding method and applied it to driving videos. Wang~\etal~\cite{wang2018video} predicted future semantic segmentation maps then translated them into images.  Other work has improved image quality using adversarial losses~\cite{lee2018savp,wang2018video,liang2017dual}, multiscale models~\cite{mathieu2015deep}, and recurrent networks with contextual aggregation~\cite{byeon2018contextvp}. These methods are complementary to ours, as our loss is architecture agnostic. Other work uses video prediction methods for model-based reinforcement learning~\cite{oh2015atari, finn2016vidpred, ha2018worldmodels}. 
Recently, Jayaraman~\etal~\cite{jayaraman2018tap} proposed time-agnostic prediction, which gives a model flexibility to predict {\em any} of the future frames in a video. Our method proposes a similar mechanism, but in space, rather than time.

Another line of work has addressed the challenges of uncertainty in video prediction through stochastic models, such as variational autoencoders (VAEs)~\cite{kingma2013vae}, that learn the full distribution of  outcomes which is then sampled~\cite{babaeizadeh2017svvp, xue2016visual, denton2018stochastic}. Notably, Denton and Fergus~\cite{denton2018stochastic} introduced a recurrent variational autoencoder that used a learned prior distribution. This work was later extended by Villegas~\etal~\cite{villegas2019high}, which introduced architectural changes and significantly increased the scale of the model, obtaining impressive results. Castrej\'{o}n~\etal~\cite{castrejon2019vrnn} also observed that high capacity models improve generation results. Other work has introduced compositional models~\cite{ye2019compositional} and sparse predictions~\cite{walker2016uncertain,hao2018controllable,ho2019sme}. Our approach is complementary to this line of work; in our experiments we obtain  benefits from using our loss in VAE-based video prediction~\cite{denton2018stochastic,villegas2019high}.

\mypar{Flow-based video prediction.}

Rather than directly outputting images, many video prediction methods instead predict optical flow for each pixel and then synthesize a result by warping the input image. In early work, Patraucean~\etal~\cite{patraucean2015spatio} predicted optical flow using a convolutional recurrent network. Liu~\etal~\cite{liu2017video} predicted a 3D space-time flow field. Gao~\etal~\cite{gao2019disentangling} regressed motion using optical flow as pseudo-ground truth, inpainted occlusions, and used semantic segmentation. Later, Wu~\etal~\cite{wu2020future} extended this approach, conditioning predictions on the trajectories of objects that their model segments and tracks. Recent work has used other motion representations, such as  factorizing a scene into stationary and moving components~\cite{denton2017disentangle,villegas2017decompose,tulyakov2018mocogan},  per-pixel kernels~\cite{finn2016unsupervised,vondrick2017generating,niklaus2017video,reda2018sdc,kong2019multigrid}, or Eulerian motion~\cite{liu2018dyan}. Work in 3D view synthesis has adopted a similar approach, known as {\em appearance flow}~\cite{zhou2016flow,park2017transformation,park2017transformation}. Since these methods can only ``copy and paste" existing content, they require special architectures to account for  disocclusion and photometric changes. For example, state-of-the-art methods~\cite{wu2020future,gao2019disentangling} have motion estimation layers, internally perform warping via spatial transformers~\cite{jaderberg2015spatial}, and use separate inpainting modules~\cite{liu2018image,yu2018inpainting} to handle disocclusion.  
Since our approach only changes the loss function, it could in principle be combined with these architectures.

\mypar{Perceptual losses.} One way of reducing blur, commonly used in video prediction~\cite{gao2019disentangling, chen2017CRN, wu2020future}, is to use perceptual losses~\cite{gatys2016image, johnson2016perceptual}. These methods exploit the invariances learned for object recognition to provide robustness to small positional errors. However, because object recognition models learn only partial invariance, when positional errors are more than a few pixels they result in the same blurring artifacts seen in simpler, pixel-based losses. In our experiments we show that this blurring can be reduced when applying our method.

\mypar{Optical flow.}
Our method uses optical flow to obtain per-pixel correspondences. To solve this task, Lucas and Kanade~\cite{lucas1981iterative,baker2004lucaskanade} make a brightness constancy assumption and solve a linearized model. Horn and Schunck~\cite{hornschunck1981hs} then proposed a smoothness prior for predicting flow. This approach was extended to use robust estimation methods~\cite{sun2010secrets,brox2004high,liu2009beyond,felzenszwalb2004efficient}. More recent methods use CNNs trained with supervised~\cite{fischer2015flownet, ilg2017flownet, ranjan2017optical, hui2018liteflownet, yang2019volumetric, sun2017pwcnet, teed2020raft} or unsupervised learning~\cite{jason2016back, ren2017unsupervised, wang2018occlusion, liu2019ddflow, jonschkowski2020matters,bian2022learning}.  Teed and Deng~\cite{teed2020raft} proposed an architecture for incrementally refining a flow field.  While these works find space-time correspondences between frames, we instead use it to find correspondences between generated and ground truth images. In this way, our work is related to methods that use optical flow for other tasks, such as matching scenes~\cite{liu2010sift}, features~\cite{xiong2021flow}, or objects~\cite{mobahi2014compositional,zhou2015flowweb,shen2020ransac}.

\mypar{Deformable matching. }
Our approach is inspired by classic work in image matching, particularly Chamfer~\cite{barrow1977parametric} and Hausdorff~\cite{huttenlocher1993comparing} matching. These methods align a template to an image before comparison,  providing robustness to small positional errors.
A similar approach has been used in image retrieval~\cite{torralba200880}, and part-based object detection~\cite{fischler1973representation,felzenszwalb2005pictorial,tezhu2012face,felzenszwalb2009object}. Single-image depth estimation has used analogous invariances to scale or space in its loss functions~\cite{eigen2014depth,fan2017point}. Like these works, we allow images to undergo deformations before comparison, but we do so for {\em synthesis} instead of matching or detection.

\mypar{Video frame interpolation.}
Frame interpolation shares many of the same challenges as video prediction, since the position and motion of objects are often uncertain. To tackle this problem, various models have been proposed that rely on optical flow~\cite{xu2019quadratic, bao2019depthaware, lee2020adacof, park2020bmbcbilateral, jiang2018super}, depth~\cite{bao2019depthaware}, or image kernels~\cite{lee2020adacof, choi2020cain}. Recently, Kalluri~\etal~\cite{kalluri2021flavr} proposed a 3D CNN for interpolation. We augment this architecture with our loss and show improvements in performance.

%% file: sections/3_method.tex
\vspace{-2mm}
\section{Correspondence-wise Image Prediction}

\input{tables/pseudocode}

Our goal is to solve image prediction tasks where there is uncertainty in the positions of objects. To address this problem, we propose a ``loss extension" that provides robustness to small positional misalignment. Given two images $\imgt$ and $\imest$, a traditional {\em pixel-wise} loss (\eg, $L_1$ distance between images) can be written as:
\vspace{-2mm}
\begin{equation}
     \loss_{P}(\im, \hat{\im}) = \frac{1}{|\mathcal{P}|}\sum_{p \in \mathcal{P}} 
     \mathcal{L}(\im_p, \hat \im_p ),
     \label{eq:pw}
     \vspace{-2mm}
\end{equation} 
where $\mathcal{L}$ is a {\em base} loss (e.g., $L_1$ for a pair of pixel intensities), $\imgt_p$ is the pixel color at location $p$ in $\imgt$, and $\mathcal{P}$ is the set of all pixel indices. 

In our approach, instead of comparing pixels at the same indices as in pixel-wise losses, we first compute pixel-to-pixel correspondences between the images, $\flowfn(\imgt, \imest)$, using optical flow. Then we compare each pixel $\imgt_p$ to its corresponding pixel $\imest_{\flowval(p)}$, where $\flowval(p)$ is the pixel in $\imest$ matched by the flow field. This loss can be written: 
\vspace{-2mm}
\begin{equation}
     \loss_{C}(\im, \hat{\im}) = \frac{1}{|\mathcal{P}|}\sum_{p \in \mathcal{P}} 
     \mathcal{L}(\im_p, \hat \im_{\flowval(p)} ),
     \label{eq:corresp}
     \vspace{-2mm}
\end{equation} 
We call the resulting loss a {\em correspondence-wise loss}. For example, when $\mathcal{L}=L_1$, we call it a correspondence-wise $L_1$ loss. The loss is illustrated in~\fig{fig:teaser}.

We first detail the implementation of the loss in~\sect{sect:regularization} and \ref{sect:implementation}, and then investigate its properties in~\sect{sect:toy_experiment}. Pseudocode for the full method, including all of the following implementation details, is provided in Alg.~\ref{alg:code}.

\input{tables/toy}

\vspace{-1mm}
\subsection{Regularization}\label{sect:regularization}
\vspace{-1mm}
\paragraph{Flow scaling.} Models that minimize~\eqn{eq:corresp} can fall into local optima during training, especially in the early stages when images are out-of-domain for the flow network. In addition, the warping process ``snaps" objects to their exact ground truth positions, making it hard for the model to infer where to place objects in the generated image.

To address these issues, we introduce a small multiplicative decay to the flow field: $\flowfn_R(\im_1, \im_2) = (1 - \epsilon) \flowfn(\im_1, \im_2)$. This reduces long-range matching while encouraging the model to place objects closer to their true locations in the target image. In each training step, a model can decrease its loss if it moves an incorrectly placed object slightly closer to its true location. We use $\epsilon = 0.1$ and call this regularization strategy {\it flow scaling}. 

\mypar{Alternative methods.} We also considered an alternative regularization strategy, inspired by Chamfer distance~\cite{barrow1977parametric} and optical flow smoothness~\cite{sun2010secrets,jonschkowski2020matters} that directly penalizes the distance that each pixel moves in the flow field. When using this scheme, our regularization term is
\vspace{-2mm}
\begin{equation}
    \mathcal{L}_{reg}(\flow) =  \lambda_1 ||\flow||^2 +  \lambda_2 \mathcal{L}_{edge}(\nabla \flow),
    \vspace{-1mm}
\end{equation}
where $\mathcal{L}_{edge}$ is the edge-aware first-order smoothness penalty of Jonschkowski~\etal~\cite{jonschkowski2020matters}, $\nabla \flow$ are the flow gradients, and $\lambda_i$ are weights. While we found this approach to be effective in some applications, we found that flow scaling generally performed better, and requires fewer hyperparameters (see \sect{sect:ablations}).

\subsection{Implementation details} \label{sect:implementation}
\vspace{4mm}

\mypar{Finding correspondences.} In order to find $\mathcal{F}(\imgt, \imest)$, we use RAFT~\cite{teed2020raft}, an optical flow network which predicts dense correspondences. In addition we have found that other models work, such as PWC-Net~\cite{sun2017pwcnet} (see Section~\ref{sect:ablations}). 

\mypar{Warp formulation.} \eqn{eq:corresp} can be evaluated by iterating over pixel locations, calculating $\mathcal{F}(p)$, taking distances, and then averaging. However, in practice, we implement our method as an image warp, followed by a pixelwise loss: 

\vspace{-1mm}\begin{equation}
    \mathcal{L}_{C}(\imgt, \imest) \approx \mathcal{L}(\imgt, \operatorname{warp} (\imest, F (\imgt, \imest ) ) ),
    \label{eq:warpdist}
\end{equation}

\noindent where $\operatorname{warp}(\imest, \flow)$ is a backward warp of $\imest$ using the deformation field $\flow$. Intuitively, the warp operation aligns pixels with their correspondences, after which we can apply an existing loss function. This formulation makes it straightforward to turn existing loss functions (e.g., perceptual losses~\cite{johnson2016perceptual} that operate on patches) into correspondence-wise losses, by warping and then applying the loss.

\mypar{Symmetry.}  Following common practice for matching-based loss functions~\cite{huttenlocher1993comparing}, we make the loss symmetric using $\mathcal{L}_{sym} = \mathcal{L}_{C}(\imgt, \imest) + \mathcal{L}_{C}(\imest, \imgt)$. This discourages models from generating superfluous content that has no correspondence in the target image.

\input{tables/ablation_figure}

\subsection{Analyzing Correspondence-wise Prediction}\label{sect:toy_experiment}

To help understand how correspondence-wise losses address the challenges of positional uncertainty, we analyze its behavior on several simplified toy prediction tasks.

\mypar{Motion uncertainty.}  
We create a simple prediction task with inherent uncertainty in position. In the example shown in~\fig{fig:toy_problem}, an object moves horizontally at unknown speed to a position uniformly distributed in a region near the image center. We ask what the optimal prediction is under various losses. For a loss $\loss$, this is the image $\overline{\im}$ that minimizes the expected loss $\Ex_{\im \sim \mathcal{D}}[\loss(\im, \overline{\im})]$, where $\mathcal{D}$ is the image distribution. We find $\overline \im$ using stochastic gradient descent.

For $\loss = L_1$, the resulting image suffers from blurring\footnote{We note that there is an analytical solution for the $L_1$ loss, namely the median, which our SGD results obtain.}. By contrast, our correspondence-based $L_1$ loss leads to a sharp prediction, with the object located in the center of the distribution. More generally, our loss tends to favor {\it crisp predictions} that commit to a single position over blurry ones. This is for two reasons: (i) blurry predictions are harder to match than sharper images via a smooth flow field, and (ii) images far from their correct position are harder to match. The same behavior occurs for the MSE loss and on other scenes~\supparxivcr{(see supplement)}{(see Figure~\ref{fig:supp_toy} in appendix)}{(see appendix)}.

\mypar{Effect of positional error.} Next, we asked how sensitive the loss is to position. For instance, could a lazy prediction method simply place objects in incorrect positions, nonetheless obtaining low loss when optical flow ``fixes" the mistake, {\em e.g.}, a video prediction model that merely repeats the last frame?  We visualize the loss as a function of an object's positional error in  \fig{fig:toy_problem}, i.e. the loss incurred if the object were predicted a given offset from its true position. To reduce the effect of the background, we averaged the results over a large number of backgrounds \supparxivcr{(see supplement for details)}{(see Section~\ref{apdx:toy} in the appendix for details)}{(see appendix for details)}. We see that our loss, in fact, {\em steadily increases with positional offset}. Moreover, the global minimum remains the same. Three reasons we found for this are: (i) Almost any incorrect prediction will have {\em occluded or extraneous content} that incurs a large loss. (ii) When flow regularization is used, the model is explicitly penalized for incorrect predictions. (iii) The implicit smoothness prior in optical flow estimation trades off reconstruction error for small, simple motions; flow methods will tend to choose matches that incur large reconstruction error over those with large flow values. 


%% file: tables/pseudocode.tex
\begin{algorithm}[t]
\small
    \caption{\small Pseudocode in a PyTorch-like style for training an image prediction method with a correspondence-wise $L_1$ loss.
    }
    \label{alg:code}
    \algcomment{\fontsize{7.2pt}{0em}\selectfont \texttt{warp}: bilinear warping with an optical flow field. \vspace{-4mm}} 

        \definecolor{codeblue}{rgb}{0.25,0.5,0.5}
        \lstset{
          backgroundcolor=\color{white},
          basicstyle=\fontsize{7.2pt}{7.2pt}\ttfamily\selectfont,
          columns=fullflexible,
          breaklines=true,
          captionpos=b,
          commentstyle=\fontsize{7.2pt}{7.2pt}\color{codeblue},
          keywordstyle=\fontsize{7.2pt}{7.2pt},
          escapechar=\&
        }
\begin{lstlisting}[language=python]
# Load a minibatch with source and target images
for (im_src, im_tgt) in loader: 
  # Predict image using a network
  im_est = predict_image(im_src)
  
  # Estimate optical flow in both directions
  F_est = optical_flow(im_tgt, im_est)
  F_tgt = optical_flow(im_est, im_tgt)
  
  # Regularize the flow
  F_est = (1 - epsilon) * F_est
  F_tgt = (1 - epsilon) * F_tgt
  
  # Warp using bilinear filtering
  warp_est = warp(im_est, F_est)
  warp_tgt = warp(im_tgt, F_tgt)
  
  # Bidirectional loss with existing loss function
  loss1 = l1_loss(im_tgt, warp_est)
  loss2 = l1_loss(im_est, warp_tgt)
  loss = loss1 + loss2
  
  loss.backward()
  
\end{lstlisting}
\end{algorithm}

%% file: tables/toy.tex
\begin{figure*}[t!]
    \centering
    \vspace{-1mm}
    \includegraphics[width=\textwidth]{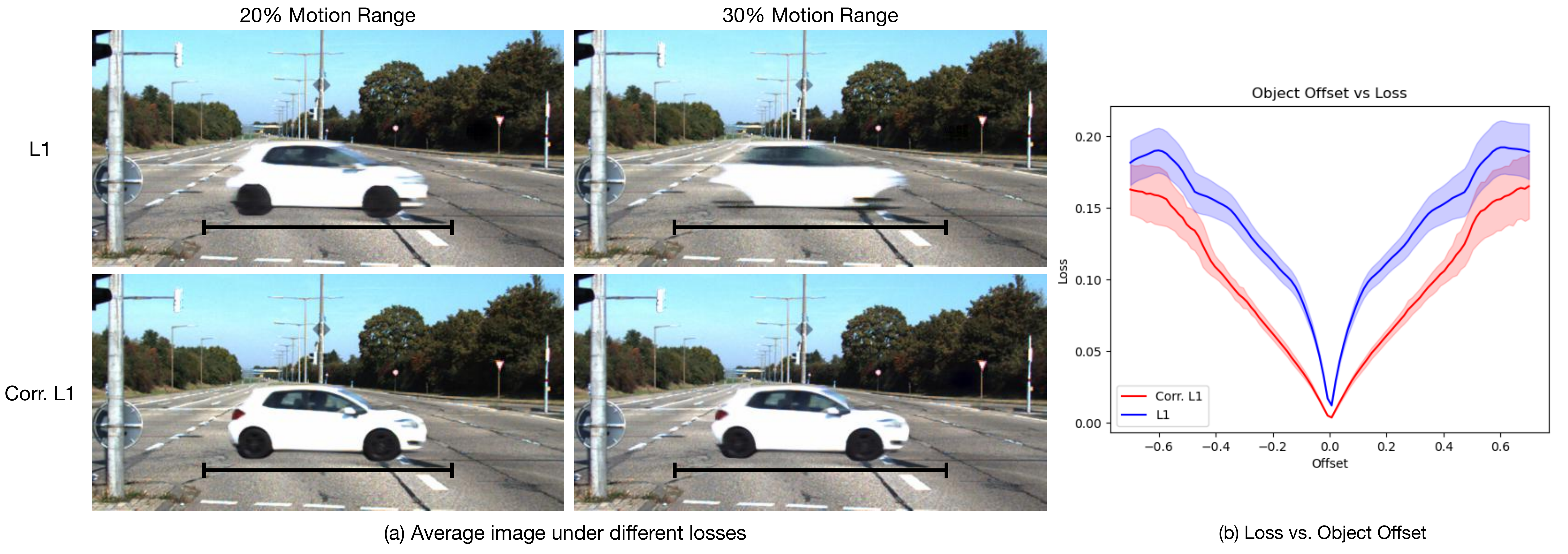}
    \vspace{-6mm}
    \caption{{\bf The effect of positional errors on different losses}. (a) We show images that minimize the expected error for a toy video prediction task. The car's location is sampled uniformly about the center at random (either $20\%$ or $30\%$ of the image width). The full range of the car's locations is indicated by the black bars. Note that the $L_1$ loss predictions are blurry, in particular for the $30\%$ setting, whereas the {\it correspondence-wise} $L_1$ loss predictions are crisp in both cases. (b) We examine how the loss changes as a function of positional error (i.e., how far the predicted object is from its true location). Correspondence-wise losses increase smoothly with magnitude of the error.} \vspace{-4mm}
    \label{fig:toy_problem}
\end{figure*}

%% file: tables/ablation_figure.tex
\begin{figure*}[]
    \centering
    \includegraphics[width=.95\linewidth]{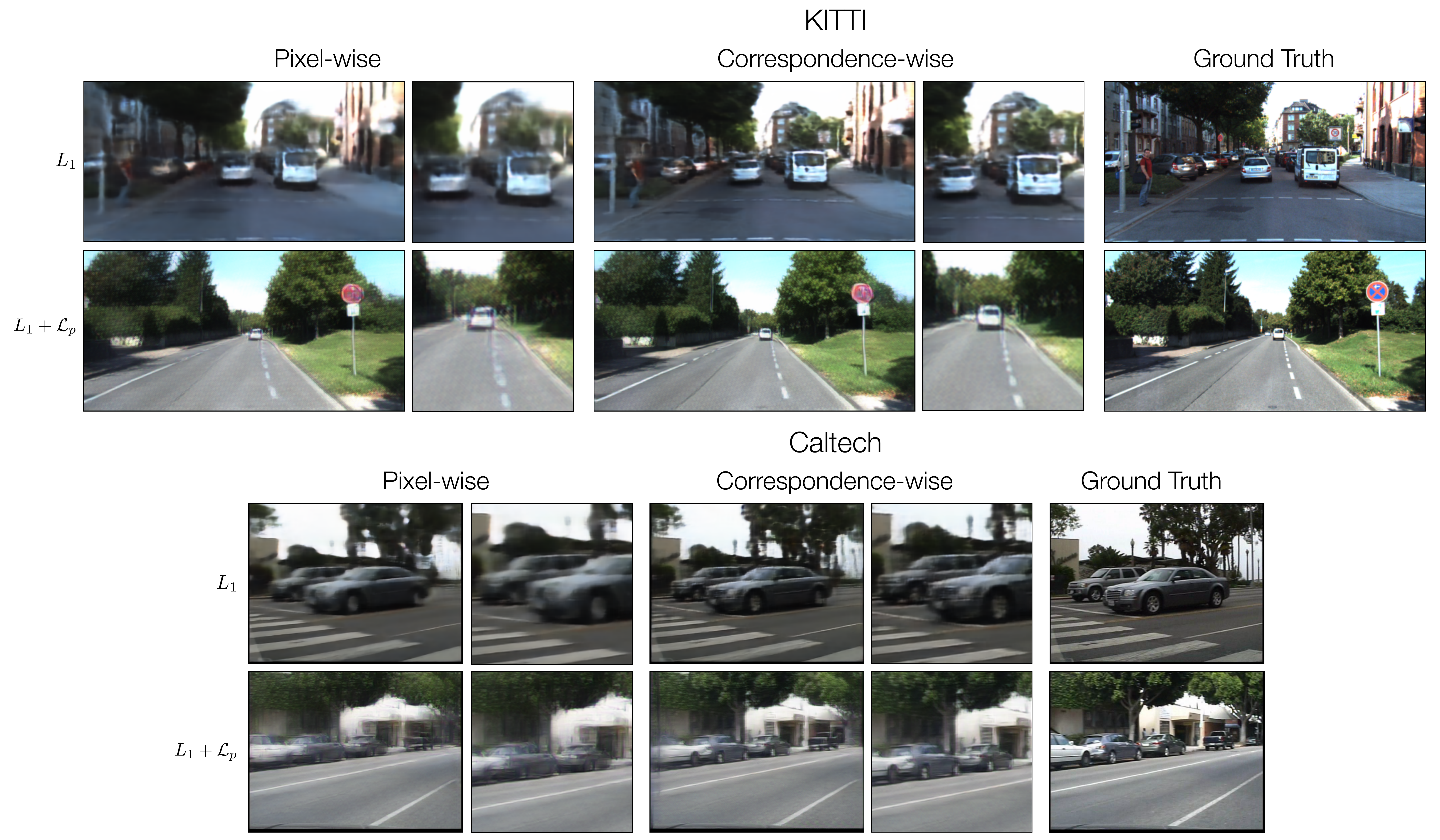}
    \vspace{-2.5mm}
    \caption{{\bf Next-frame prediction with and without a correspondence-wise loss.} We show both $L_1 + \mathcal{L}_p$ and $L_1$ results, indicated by the left most column, on KITTI and Caltech. We highlight a noteworthy part of each result. Using a correspondence-wise loss results in sharper details (e.g., crisper road lines) and more robustness to large motions.}
    \label{fig:qualitative_ablation}
    \vspace{-5pt}
\end{figure*}

%% file: sections/4_experiments.tex
\vspace{-2mm}

\input{tables/ablations_cvpr_small}

\section{Results}

Our goal is to understand how correspondence-wise losses differ from their pixelwise counterparts, and to evaluate their effectiveness. To do this we perform experiments on video prediction, as well as frame interpolation. We ablate pixelwise and correspondence-wise losses across various tasks, datasets, architectures, and metrics. In addition, we compare models trained with our loss to the state-of-the-art methods.

\vspace{-2mm}
\subsection{Video Prediction}

\paragraph{Models.} We consider both deterministic and stochastic video prediction architectures. For simplicity, our deterministic model is based on the widely-used residual network~\cite{johnson2016perceptual,he2016deep} from Wang~\etal~\cite{wang2017pix2pixhd}, except we replace 2D convolutions by 3D convolutions~\cite{tran20173dconv, carreira2017I3D} to process temporal information, and we replace transposed convolutions with upsampling followed by 2D convolutions to avoid checkerboard artifacts \supparxivcr{(see supplement for full network details)}{(see Section~\ref{apdx:training} in the appendix for full network architecture)}{(see appendix for full network architecture)}. Multi-frame prediction is performed by recursively feeding output images back into the model, as in~\cite{lotter2016predictive, gao2019disentangling, wu2020future}.

The stochastic model we use is the {\bf SVG} model introduced by Denton and Fergus~\cite{denton2018stochastic}, a CNN with a learned prior and LSTM layers. In addition, we adopt extensions to the SVG model proposed by Villegas~\etal~\cite{villegas2019high}, resulting in a model we call {\bf SVG++}. This is a large-scale model that offers architectural improvements over SVG, and which obtains strong performance on the KITTI dataset. Following Villegas \etal we modify SVG by adding convolutional LSTM layers~\cite{shi2015convolutional}, although we keep the $L_2$ loss from~\cite{denton2018stochastic} \supparxivcr{(see supplement for details)}{(see Section~\ref{apdx:training} in the appendix for more details)}{(see appendix for details)}. Since our goal is to understand the influence of a correspondence-wise loss, rather than to obtain state-of-the-art performance, we base our model on a medium-sized variant with hyperparameters $K=2$ and $M = 2$~\cite{villegas2019high}, which obtains strong performance yet is trainable on ordinary multi-GPU computing infrastructure.

For optical flow estimation, we use RAFT~\cite{teed2020raft}, with the publicly available checkpoint that was trained on Flying Chairs~\cite{dosovitskiy2015flying} and FlyingThings~\cite{flyingthingsMIFDB16}. While our results may be improved by using a version trained on driving videos from KITTI~\cite{geiger2013IJRR}, we choose not to do so in our  experiments to avoid the use of domain-specific supervision. 

\mypar{Losses.} We use three base losses: {\bf 1)} the $L_1$ loss, {\bf 2)} the MSE loss ($L_2^2$), and {\bf 3)} the $L_1$ loss in equal weight with a perceptual loss, $\mathcal{L}_p$, ($L_1 + \mathcal{L}_p$) that uses VGG-19 pretrained with ImageNet features~\cite{johnson2016perceptual,russakovsky2015imagenet,simonyan2014very}.

\mypar{Metrics.} To evaluate predictions, we use SSIM, LPIPS, and a two alternative forced choice human study \supparxivcr{(see supplement for details)}{(see Section~\ref{apdx:amt} in the appendix for more details)}{(see appendix for details)}.

\mypar{Datasets.} We evaluate on three standard video prediction datasets: KITTI~\cite{geiger2013IJRR}, Caltech Pedestrian~\cite{dollarCVPR09peds}, and Cityscapes~\cite{Cordts2016Cityscapes} \supparxivcr{(see supplement for details)}{(see Section~\ref{apdx:training} in the appendix for more details)}{(see appendix for details)}.
\vspace{-2mm}
\subsubsection{Pixelwise vs. Correspondence-wise losses}

To understand the effect of our extension of pixelwise losses, we train video prediction models to predict three future frames from three previous frames with both losses, using three separate base losses: $L_1$, $L_2^2$ (MSE), and $L_1 + \mathcal{L}_p$. We evaluate at a resolution of $512 \times 256$ on the KITTI dataset, $384 \times 288$ on Caltech, and $512 \times 256$ of Cityscapes, all with three frames of input. For consistency with the KITTI experiments, we sample the Caltech dataset at 10 Hz, as in Lotter~\etal~\cite{lotter2016predictive}. In addition, for the $L_1 + \mathcal{L}_p$ loss, we warm start with a pixelwise loss for an epoch, which significantly improves convergence.

The results can be found in~\tbl{tbl:ablations}. The correspondence-wise variants of the loss outperform their pixelwise counterparts on almost all metrics, datasets, and losses. In addition, we see significant improvements in the qualitative results across all scenarios, as seen in Figure~\ref{fig:qualitative_ablation}.

\vspace{-4mm}
\subsubsection{Comparison to State-of-the-Art Methods}

\input{tables/sota}

We follow the evaluation protocol of Wu~\etal~\cite{wu2020future}, using the KITTI and Cityscapes datasets, and compare with a number of recent video prediction methods: {\bf Voxel Flow}~\cite{liu2017video}, a motion-synthesis method based on 3D space-time flow, {\bf MCnet}~\cite{villegas2017decompose}, a convolutional LSTM model that decomposes stationary and moving components, {\bf Vid2Vid}~\cite{wang2018video}, a two-stage method that first synthesizes semantic masks and then translates the masks to real images, and {\bf OMP}~\cite{wu2020future}, a state-of-the-art method that combines copy-and-paste prediction with inpainting, object tracking, occlusion estimation, and adversarial training. We also show results with {\bf PredNet}~\cite{lotter2016predictive}, a convolutional pixel-based architecture inspired by predictive coding. We note that these architectures are specialized to the video prediction task, and can be relatively complex. For example, OMP uses off-the-shelf instance segmentation and semantic segmentation networks~\cite{xiong2019upsnet,zhu2018improving}, an inpainting network~\cite{yu2018inpainting}, and a background-prediction network. It also takes optical flow as input~\cite{sun2017pwcnet}, tracks objects, and uses adversarial training. By contrast, we are interested in seeing how well a simple image prediction network can do when using our loss. 

Following Wu~\etal~\cite{wu2020future}, we use $832 \times 256 $ images on KITTI and $1024 \times 512$ on Cityscapes, using the same train-test split and data augmentation, and condition our models on four frames of an input video and predict five future frames on KITTI and 10 on Cityscapes. We use our ResNet variant with an $L_1 + \mathcal{L}_p$ correspondence-wise loss.

Despite our method's simplicity, we found that it significantly outperformed previous methods with complex flow-based architectures on both metrics (\tbl{tbl:sota}). Our simple architecture, trained with a correspondence-wise loss, obtained higher scores on both SSIM and LPIPS consistently across all time steps.

In~\fig{fig:qualitative_omp}, we show three informative qualitative results generated by our model, as compared to frames generated by OMP. We have highlighted challenging regions in each video. In contrast to our model, OMP works by deforming an input image using a predicted optical flow field, resulting in warping artifacts when there are errors. Here, OMP suffers when there are small objects undergoing large motions (\eg, the poles in the first video) and irregular geometry (\eg, second video). Interestingly, both models produce errors in disoccluded regions (\eg, car in first video).

\vspace{-4mm}
\subsubsection{Additional Ablations}\label{sect:ablations}

We present additional ablations on network architecture, regularization scheme, and flow method for the video prediction task.

\mypar{Stochastic models.} We demonstrate the versatility of our loss by evaluating its performance with stochastic video prediction architectures on KITTI in \tbl{tbl:svg}. Evaluation metrics on both the SVG and SVG++ significantly improve when we use a correspondence-wise reconstruction loss. 

\input{tables/svg}
\input{tables/ablations_reg}
\input{tables/ablations_flow}

\mypar{Effect of regularization.} In Section~\ref{sect:regularization} we introduced two methods of regularization. \tbl{tbl:ablations_reg} evaluates these two approaches. The {\em magnitude penalty} approach outperforms the {\em scaling} approach in some circumstances. However, we found that having to tune each of its weighting parameters was challenging, and that it often had poor convergence, which we addressed by training with a warm start from a base loss for 5 epochs.

\mypar{Quality of flow method.} In Table~\ref{tbl:ablations_flow} we test the influence of the quality of the flow method by comparing RAFT~\cite{teed2020raft} and PWC-Net~\cite{sun2017pwcnet}. In addition, we simulated less powerful flow methods by simply fitting a homography or an affine transformation to the predicted RAFT flow, and using these fitted transforms as our flow. We found that using a correspondence-wise loss with better quality flow methods produces better downstream image generation performance.

\input{tables/frame_interp_noavg}
\input{tables/frame_interp_qualitative}

\subsection{Video Frame Interpolation}\label{sect:frame_interp}

To further evaluate our loss's ability to handle spatial uncertainty, we conduct frame interpolation experiments. Given two frames of context, the goal is to predict the intermediate frame. We use the FLAVR architecture~\cite{kalluri2021flavr} with both an $L_1$ loss and an $L_1$ correspondence-wise loss and we evaluate using SSIM, PSNR, and LPIPS on the Vimeo-90K septuplet dataset~\cite{xue2019video}. These results, along with various two-frame baselines and a FLAVR model trained on four frames, are presented in~\tbl{tbl:frames}.

We trained our model using the publicly released code from FLAVR, replacing their loss with our correspondence-wise loss. To be more consistent with the baselines and to study a scenario with more uncertainty, we use two context frames instead of four. Specifically we use the two middle frames, closest to the ground truth, in the septuplets. We train for 120 epochs, use a batch size of 32, and learning rate of 0.0002 with Adam~\cite{kingma2014adam}. These hyperparameters are the same for all FLAVR models.

The FLAVR model trained with our correspondence-wise $L_1$ loss outperforms  the model with only $L_1$ loss on LPIPS and PSNR, ties on SSIM, and leads to crisper interpolations (Fig.~\ref{fig:qualitative_frames}). In addition, the model outperforms all but one baseline and is even competitive with the FLAVR model trained on four frames.

%% file: tables/ablations_cvpr_small.tex
\begin{table}[t!]
    \begin{center}
    \setlength\tabcolsep{3pt}
    \resizebox{\linewidth}{!}{
    \begin{tabular}{ccccccccc}
    \toprule
    & & \multicolumn{3}{c}{KITTI} & \multicolumn{2}{c}{Cityscapes} & \multicolumn{2}{c}{Caltech} \\
    \cmidrule(lr){3-5} \cmidrule(lr){6-7} \cmidrule(lr){8-9}
    Base Loss & Corr. & SSIM $\uparrow$ & LPIPS $\downarrow$ & 2AFC $\uparrow$ & SSIM $\uparrow$ & LPIPS $\downarrow$ & SSIM $\uparrow$ & LPIPS $\downarrow$ \\
    \midrule
    $L_1$ & - & 0.563 & 0.438 & 13.52 & {\bf 0.820 } & 0.231 & 0.733 & 0.250 \\
    $L_1$ & \checkmark & {\bf 0.586 } & {\bf 0.359 } & {\bf 14.25} & 0.819 & {\bf 0.198 } & {\bf 0.734 } & {\bf 0.216 } \\
    \cmidrule(lr){1-9}
    $L_2^2$ & - & 0.544 & 0.499 & 13.36 & 0.801 & 0.291 & 0.701 & 0.330 \\
    $L_2^2$ & \checkmark & {\bf 0.563 } & {\bf 0.403 } & {\bf 15.16} & {\bf 0.812 } & {\bf 0.212 } & {\bf 0.707 } & {\bf 0.249 } \\
    \cmidrule(lr){1-9}
    $L_1+\mathcal{L}_p$ & - & 0.545 & 0.213 & 14.46 & {\bf 0.816 } & 0.092 & {\bf 0.717 } & {\bf 0.139 } \\
    $L_1+\mathcal{L}_p$ & \checkmark & {\bf 0.548 } & {\bf 0.191 } & {\bf 20.19} & 0.810 & {\bf 0.090 } & 0.702 & 0.141 \\
    \bottomrule
    \end{tabular}
   }
    \end{center}
    \vspace{-15pt}

\caption{{\bf Pixel-wise vs. correspondence-wise losses.} We compare correspondence-wise losses to their pixelwise counter parts, showing its efficacy on KITTI, Cityscapes, and Caltech. The second column indicates the usage of a correspondence-wise loss. 2AFC is the rate at which humans chose the generated video over the true video in a real-or-fake study.\vspace{-6mm} }
\label{tbl:ablations}
\end{table}

%% file: tables/sota.tex
\begin{figure*}[t]
    \centering
    \includegraphics[width=.95\linewidth]{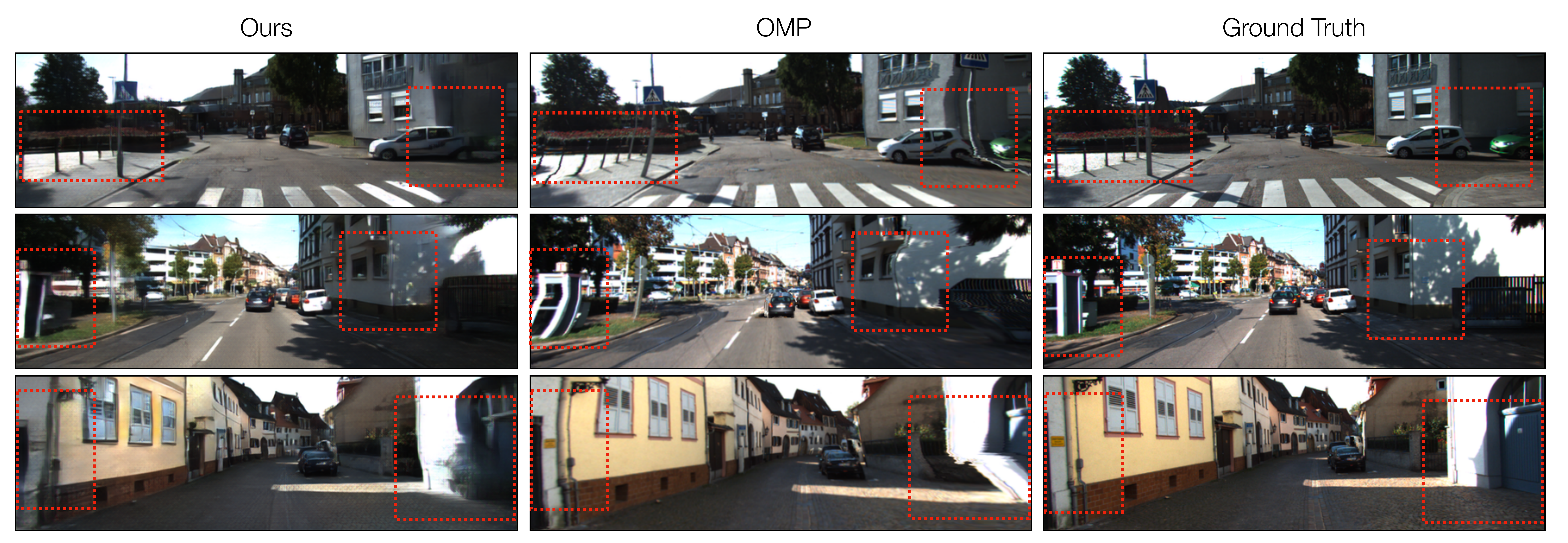} \vspace{-3mm}
    \caption{{\bf Comparison with state-of-the-art multi-frame video prediction.} We show the result of our method and OMP~\cite{wu2020future} on predicting frame $t = 5$. Areas marked with a box are challenging examples. }\vspace{-2mm}
    \label{fig:qualitative_omp}
\end{figure*}

\begin{table*}
    \begin{center}
    \resizebox{.8\linewidth}{!}{%
    \setlength\tabcolsep{2pt}
    \begin{tabular}{lcccccc|cccccc}
    \toprule
    &  \multicolumn{6}{c|}{KITTI} & \multicolumn{6}{c}{Cityscapes} \\
    &  \multicolumn{2}{c}{Next Frame} & \multicolumn{2}{c}{Next 3 Frames} & \multicolumn{2}{c|}{Next 5 Frames} & \multicolumn{2}{c}{Next Frame} & \multicolumn{2}{c}{Next 5 Frames} & \multicolumn{2}{c}{Next 10 Frames} \\
    Model & SSIM $\uparrow$ & LPIPS$\downarrow$ & SSIM $\uparrow$ & LPIPS$\downarrow$ & SSIM $\uparrow$ & LPIPS$\downarrow$ & SSIM $\uparrow$ & LPIPS$\downarrow$ & SSIM $\uparrow$ & LPIPS$\downarrow$ & SSIM $\uparrow$ & LPIPS$\downarrow$ \\
    \midrule
    PredNet~\cite{lotter2016predictive} & 0.563 & 0.553 & 0.514 & 0.586 & 0.475 & 0.629 & 0.840 & 0.260 & 0.752 & 0.360 & 0.663 & 0.522 \\
    MCNET~\cite{villegas2017decompose} & 0.753 & 0.240 & 0.635 & 0.317 & 0.554 & 0.373 & 0.897 & 0.189 & 0.706 & 0.373 & 0.597 & 0.451 \\
    Voxel Flow~\cite{liu2017video} & 0.539 & 0.324 & 0.469 & 0.374 & 0.426 & 0.415 & 0.839 & 0.174 & 0.711 & 0.288 & 0.634 & 0.366 \\
    Vid2Vid~\cite{wang2018video} & - & - & - & - & - & - & 0.882 & 0.106 & 0.751 & 0.201 & 0.669 & 0.271 \\
    OMP~\cite{wu2020future} &  0.792 & 0.185 & 0.676 & 0.246 & 0.607 & 0.304 & 0.891 & {\bf 0.085} & 0.757 & 0.165 & 0.674 & 0.233 \\
    \midrule
    Ours & {\bf 0.820} & {\bf 0.172} & {\bf 0.730} & {\bf 0.220} & {\bf 0.667} & {\bf 0.259} & {\bf 0.928} & {\bf 0.085} & {\bf 0.839} & {\bf 0.150} & {\bf 0.751} & {\bf 0.217} \\
    \bottomrule
    \end{tabular}%
    }
    \end{center}
    
    \vspace{-6mm}
    \caption{{\bf Comparisons with other methods.} We compare with state-of-the-art methods on multi-frame KITTI and Cityscapes video prediction, as well PredNet, an older non-flow based method. Note that in keeping with previous works, we present the {\it average} of the first $ N $ frames and we use the linear variant of LPIPS. We take baseline numbers directly from Wu~\etal~\cite{wu2020future}. \vspace{-3mm}}
    \label{tbl:sota}
\end{table*}

%% file: tables/svg.tex
\begin{table}
    \begin{center}
    \resizebox{\linewidth}{!}{%
    \setlength\tabcolsep{1.5pt}
    \begin{tabular}{lccccccc}
    \toprule
    & & \multicolumn{2}{c}{Next Frame} & \multicolumn{2}{c}{Next 5 Frames} & \multicolumn{2}{c}{Next 10 Frames}\\
    Model & Corr. & SSIM $\uparrow$ & LPIPS $\downarrow$ & SSIM $\uparrow$ & LPIPS $\downarrow$ & SSIM $\uparrow$ & LPIPS $\downarrow$ \\
    \midrule
    \multirow{2}{*}{SVG~\cite{denton2018stochastic}}
    & - & 0.389 & 0.478 & 0.342 & 0.509 & 0.312 & 0.527 \\
    & \checkmark & {\bf 0.400} & {\bf 0.389} & {\bf 0.348} & {\bf 0.382} &  {\bf 0.313} & {\bf 0.386} \\
    \cmidrule(lr){1-8}
    \multirow{2}{*}{SVG++~\cite{villegas2019high}} 
    & - & 0.848 & 0.079 & 0.626 & 0.196 & 0.489 & 0.287 \\
    & \checkmark & {\bf 0.849} & {\bf 0.072} & {\bf 0.628} & {\bf 0.186} & {\bf 0.490} & {\bf 0.276} \\
    \bottomrule
    \end{tabular}%
    }
    \end{center}
    
    \vspace{-6mm}
    \caption{{\bf Stochastic model evaluation.} 
    We show results with and without our correspondence-wise loss (``Corr.") while using stochastic video prediction architectures.}
    \vspace{-4mm}
    \label{tbl:svg}
\end{table}

%% file: tables/ablations_reg.tex
\begin{table}
    \begin{center}
    \setlength\tabcolsep{2pt}
    \resizebox{\linewidth}{!}{
    \begin{tabular}{cccccccc}
    \toprule
    & & \multicolumn{2}{c}{KITTI} & \multicolumn{2}{c}{Caltech} & \multicolumn{2}{c}{Cityscapes} \\
    \cmidrule(lr){3-4} \cmidrule(lr){5-6} \cmidrule(lr){7-8}
    Base Loss & Reg & SSIM $\uparrow$ & LPIPS $\downarrow$ & SSIM $\uparrow$ & LPIPS $\downarrow$ & SSIM $\uparrow$ & LPIPS $\downarrow$ \\
    \midrule
    $L_1$ & Scale & {\bf 0.586} & {\bf 0.359} & {\bf 0.734} & {\bf 0.216} & 0.819 & {\bf 0.198} \\
    $L_1$ & Mag. & 0.573 & 0.407 & 0.723 & 0.249 & {\bf 0.826} & 0.227 \\
    \cmidrule(lr){1-8}
    $L_1+\mathcal{L}_p$ & Scale & {\bf 0.548} & {\bf 0.191} & 0.702 & {\bf 0.141} & 0.810 & {\bf 0.090} \\
    $L_1+\mathcal{L}_p$ & Mag. & {\bf 0.548} & 0.206 & {\bf 0.707} & 0.149 & {\bf 0.823} & 0.098 \\
    \bottomrule
    \end{tabular}
    }
    \end{center}
\vspace{-6mm}
\caption{{\bf Regularization for correspondence-wise losses.} We ablate the flow scaling and flow magnitude penalty regularization methods, ``Scale" and ``Mag." in the table respectively. These experiments correspond to the ones in~\tbl{tbl:ablations}.}
\vspace{-4mm}
\label{tbl:ablations_reg}
\end{table}

%% file: tables/ablations_flow.tex
\begin{table}[h]
    \begin{center}
    \setlength\tabcolsep{3pt}
    \resizebox{.7\linewidth}{!}{
    \begin{tabular}{lccccc}
    \toprule
    Metric & Pixelwise & Affine & Hom. & PWC-net & RAFT \\
    \midrule
    LPIPS $\downarrow$ & 0.438 & 0.393 & 0.393 & 0.364 & {\bf 0.359} \\ 
    \rule{0pt}{2.6ex} SSIM $\uparrow$ & 0.563 & 0.568 & 0.569 & {\bf 0.586} & {\bf 0.586}\\
    \bottomrule
    \end{tabular}
    }
    \end{center}
    
    \vspace{-5.5mm}

    \caption{{\bf Influence of flow method.} We show image quality as a function of flow method with an $L_1$ base loss on KITTI (the setup in the first two rows of Table~\ref{tbl:ablations}).}
    \label{tbl:ablations_flow}
    \vspace{-4mm}
\end{table}

%% file: tables/frame_interp_noavg.tex
\begin{table}
    \begin{center} \small
    \setlength\tabcolsep{3pt}
    \resizebox{.95\linewidth}{!}{
    \begin{tabular}{lcccc}
    \toprule
    Method & Frames & LPIPS ($\downarrow$) & PSNR ($\uparrow$) & SSIM ($\uparrow$) \\
    \midrule
    DVF ~\cite{liu2017dvf} & 2 & - & 27.27 & 0.893\\
    SuperSloMo ~\cite{jiang2018super} & 2 & - & 32.90 & 0.957\\
    SepConv ~\cite{niklaus2017video} & 2 & - & 33.60 & 0.944\\
    CAIN ~\cite{choi2020cain} & 2 & - & 33.93 & 0.964\\
    BMBC ~\cite{park2020bmbcbilateral} & 2 & - & 34.76 & 0.965\\
    AdaCoF ~\cite{lee2020adacof} & 2 & - & 35.40 & 0.971\\
    FLAVR ~\cite{kalluri2021flavr} & 4 & 0.0248 & 36.30 & 0.975\\
    \midrule
    FLAVR ~\cite{kalluri2021flavr} & 2 & 0.0297 & 34.96 & 0.970\\
    FLAVR + Ours & 2 & 0.0268 & 35.13 & 0.970 \\
    \bottomrule
    \end{tabular}
    }
    \end{center}
    
    \vspace{-6mm}
    \caption{{\bf Frame interpolation evaluation.}  We replace the $L_1$ loss in FLAVR with a correspondence-wise $L_1$ loss. We report baseline PSNR and SSIM numbers from Kalluri \etal~\cite{kalluri2021flavr}.}
    \label{tbl:frames}
    \vspace{-4mm}
\end{table}


%% file: tables/frame_interp_qualitative.tex
\begin{figure*}[]
    \centering
    \includegraphics[width=.95\linewidth]{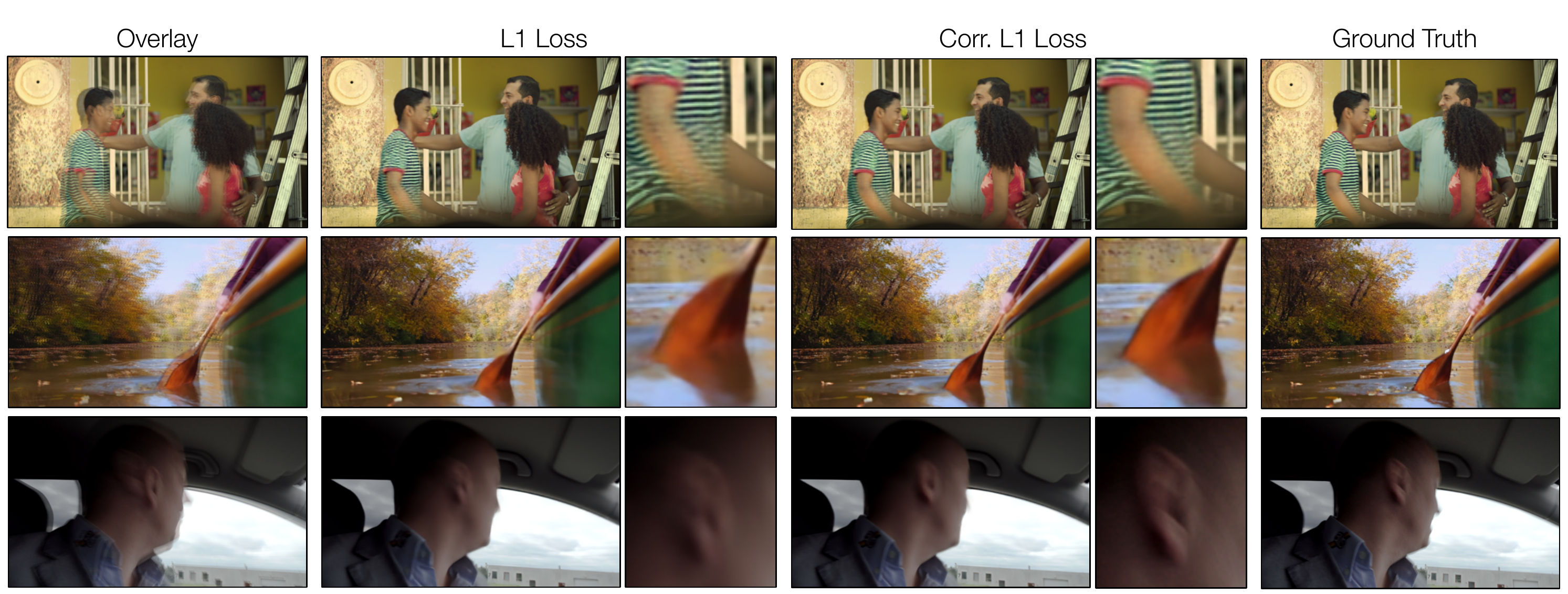}
    \vspace{-2.5mm}
    \caption{{\bf Frame interpolation qualitative results.} We visualize frame interpolation results from Vimeo-90K for a FLAVR model trained on an $L_1$ loss and a correspondence-wise $L_1$ loss. The leftmost column shows the two context frames, overlaid on top of each other. We highlight a noteworthy region of each result. \vspace{-2mm}
    }
    \label{fig:qualitative_frames}
    \vspace{-5pt}
\end{figure*}

%% file: sections/5_discussion.tex


\section{Discussion}
\vspace{2mm}

\mypar{Limitations.} One drawback of our method is that it adds overhead at training time, since it requires evaluating an optical flow network twice per example. In our ablation experiments, corresponding to Table~\ref{tbl:ablations}, we found an increase of approximately 30\%-50\% wall time. However, test time inference speed is unaffected as our method is a loss function. Additionally, our method requires the optical flow model to successfully match the predicted and ground truth images, which may not be possible in some tasks. 

\mypar{Conclusion.} In this paper we have proposed a {\em correspondence-wise} loss for image generation, and applied it to the task of video prediction and frame interpolation. We show through extensive ablations that our loss improves image quality over pixelwise losses on various metrics, architectures, datasets, and tasks. Despite the simplicity of our approach, it outperforms several recent, highly engineered video prediction methods when used with simple off-the-shelf architectures.

We see our work opening two directions. The first is to unify ``copy-and-paste" synthesis methods~\cite{zhou2016flow,patraucean2015spatio} and correspondence-wise methods by designing architectures that can obtain the benefits of both techniques. The second is to use  correspondence-wise image prediction in other tasks where positional uncertainty currently leads to blurry results, such as viewpoint synthesis and image translation. 


\mypar{Acknowledgements.} We thank Ruben Villegas, Yue Wu, Hang Gao, and Dinesh Jayaraman for helpful discussions. This material is based upon work supported by the National Science Foundation Graduate Research Fellowship under Grant No. 1841052.

%% file: sections/supplementary.tex

\appendix
\renewcommand{\thesection}{A\arabic{section}}
\renewcommand{\thefigure}{A\arabic{figure}}
\renewcommand{\thetable}{A\arabic{table}}
\setcounter{section}{0}
\setcounter{figure}{0}
\setcounter{table}{0}

\section{Toy Image Prediction Details}\label{apdx:toy}

\paragraph{Additional Results.} We conducted additional experiments using the toy image prediction task considered in~\sect{sect:toy_experiment}. In~\fig{fig:supp_toy} we conduct the same experiment with various combinations of  backgrounds and objects and provide results of the experiment with $L_2^2$ loss. We see that a correspondence-wise loss consistently gives sharp reconstructions. 

\mypar{Experimental Details.} The model used for reconstruction is our baseline model (described in Section~\ref{apdx:training}), with a fixed random Gaussian image as input. We train for 10,000 steps with a learning rate of $2\times10^{-4}$ with Adam~\cite{kingma2014adam}, which we find is more than enough for convergence. Jittering of the object is performed by compositing the object onto a fixed background with a horizontal offset from the center drawn from a uniform distribution.

\mypar{Loss Plot Details.} In Figure~\ref{fig:toy_problem} we plot the positional error of an object (a car) against loss. To reduce the variation in the loss landscape due to the choice of background we average the loss landscapes over 32 different background images randomly drawn from the KITTI train set. The shaded regions denote a 95\% confidence interval.

\input{tables/toy_extra}

\section{Training and Architecture}\label{apdx:training}

\paragraph{Video Prediction Architecture.} A table describing our architecture can be found in Table~\ref{tbl:architecture}. The video prediction model is adapted from Wang~\etal in Pix2PixHD~\cite{wang2017pix2pixhd}, which itself was based on the ResNet architecture of Johnson~\etal~\cite{johnson2016perceptual}. We do not use the adversarial loss. This architecture consists of a convolutional downsampling front-end, a set of residual blocks, and a transposed convolutional upsampling back-end. For video prediction tasks, we replace some 2D convolutions with 3D space-time convolutions to exploit temporal information~\cite{tran20173dconv, carreira2017I3D}. In addition, we replace instance normalization with batch normalization, and transposed convolutions with nearest neighbor upsampling followed by a convolutional layer. Using nearest neighbor upsampling reduces checkerboard artifacts, but interestingly we found that using the perceptual loss reintroduces faint artifacts. We suspect this is due to artifacts in the gradients of the underlying VGG19 network.

The initial downsampling block contains three stride-two convolutions in the spatial dimensions, halving the image size at each layer while keeping the temporal dimension fixed, ending with a feature map that is $ 8 \times $ smaller spatially, with 512 feature dimensions. This is then followed by two 3D ResNet blocks. Four 2D ResNet blocks are then applied to the last feature map in the temporal dimension. All other features are discarded. Finally, nearest neighbor upsampling interlaced with convolutional layers process the feature maps back to the original image size and a single convolutional layer is applied to reduce the feature dimension to three RGB channels. We use batch normalization for all the normalization layers and ReLU non-linearities everywhere except for the final layer, which uses a $\tanh$ layer.

\input{tables/architecture}

\mypar{Training Details.} For video prediction we use the recursive training strategy as proposed by Lotter~\cite{lotter2016predictive}, in which the outputs of the network are used as context to predict future frames. The loss is then backpropagated through the entire chain of recursive function applications and gradients are accumulated. We use Adam~\cite{kingma2014adam} with learning rate $ 2 \times 10^{-4} $ for all experiments. We use a batch size of 3 for all experiments due to GPU memory limitations. Weights are initialized from a Gaussian distribution with mean 0 and standard deviation $ 0.02 $. We train for 20 epochs for a total of about $2.5\times 10^5$ iterations.

\mypar{Dataset Details} For the KITTI and Caltech dataset we use the same train-test splits as in Lotter~\etal~\cite{lotter2016predictive}, and we sample Caltech at 10Hz, also as in~\cite{lotter2016predictive}. For Cityscapes we use the dataset containing 30 frame snippets sampled at 17 Hz, and use the given train test split.

\mypar{SVG and SVG++}
To implement SVG we use the publicly available source code on Github. We implement SVG++ from the SVG source code with guidance from the authors. For SVG++ we use the $L_2$ loss instead of the $L_1$ loss as mentioned in the paper, so that the primary difference between the two is the architecture (rather than the loss function). For SVG, we use an embedding dimension of 128, a latent dimension of 24, 256 units for all RNNs, and a learning rate of $8\times 10^{-4}$ with Adam. For SVG++ we use an embedding dimension of 512, a latent dimension of 128, 512 units for all RNNs, and a learning rate of $1\times 10^{-4}$ with Adam.

\input{tables/frame_interp_skip}

\section{Design Decisions}
\vspace{4mm}

\mypar{Regularization} For the flow magnitude regularization strategy, discussed in~\sect{sect:regularization}, we tuned the hyperparameters by random search evaluated on the KITTI next frame prediction task. We found that for the $L_1$ base loss, $\lambda_1 = 9.0$ and $\lambda_2 = 4.5$ worked best. In addition, we replace the edge aware smoothing, $\mathcal{L}_{edge}(\nabla \flow)$, with the norm of the flow gradient, $|\nabla \flow|$. For the $L_1+\mathcal{L}_p$ base loss, we use $\lambda_1 = 6.5$ and $\lambda_2 = 6.5$ with the first order edge-aware smoothing penalty. All flow magnitude regularization models were warm-started for five epochs.

For the flow scaling regularization strategy, we present a version of the plot in ~\fig{fig:toy_problem}(b) in ~\fig{fig:flow_scaling}, where we ablate out flow scaling. Qualitatively, we find that flow scaling results in a loss that is more directly proportional to the object offset.

\input{tables/toy_loss}

\mypar{Symmetrization} In theory our loss can be used asymmetrically in which only the generated image is matched to the ground truth image, or vice versa. We found that this approach produces results that are often quantitatively similar to the symmetrized loss, but also sometimes results in very obvious shifts in the background. This is especially prominent in the ``toy" experiments in which the background is completely static. We hypothesize that this is because the asymmetric version is not constrained enough. For example, if we only warp the ground truth image into the generated image then the generated image can contain extraneous pixels, as long as it also contains the generated image (perhaps resized slightly). For this reason we use the symmetric loss in all our experiments.

\section{Frame Interpolation Details}\label{apdx:fi}

\paragraph{Training Details.}
For our frame interpolation experiments we utilize the code from Kalluri \etal\cite{kalluri2021flavr} but set the model to have two context frames instead of four. Specifically we use the two middle frames, closest to the ground truth in the septuplets. We train for 120 epochs, use a batch size of 32, and learning rate of 0.0002 with Adam~\cite{kingma2014adam}. These hyper parameters are the same for all models.

\mypar{Longer-range Interpolation.}
We performed additional experiments for the frame interpolation task in~\sect{sect:frame_interp} by modifying which two context frames are utilized. Alternatively, we can select the first and last frames in the septuplet, instead of the middle two. This increases the time between the two input context frames by a factor of three, resulting in significantly more uncertainty for the model. 

The results from this configuration are shown in Table \ref{tbl:frames_skip}. We find that the FLAVR model trained with our correspondence-wise loss is nearly identical to the baseline FLAVR model in terms of PSNR and SSIM, while having significantly lower LPIPS. Similar to the results in~\sect{sect:frame_interp}, our loss results in much crisper interpolations. These additional results are especially promising for low framerate settings, or videos with fast movement.

\section{Model Predictive Control}\label{sect:mpc}
We consider applying our loss function to simple visual model predictive control tasks (MPC from pixels). These methods require measuring the distance in state space given an estimated future {\it image} observation and the current image observation. Simple pixel-wise distances, such as the commonly-used $L_1$ loss, often fail to convey distance in state space. For example, in~\fig{fig:mpc_ant} the state distance (delta between poses) cannot be accurately inferred by taking an $L_1$ loss between the two images. Due to these limitations, many approaches contain significant handcrafted components (e.g. requiring human in the loop~\cite{ebert2018vf}). We compare our correspondence-wise $L_1$ loss function to a pixel-wise $L_1$ loss in visual MPC, while avoiding the complexity of other costs.

We test MPC on the Cartpole, Ant, and Humanoid environments with an oracle action-conditioned predictor. The goal for all three tasks is to attain the same state as a randomly sampled goal. We take 100 rollouts and report the average distance from the goal in state space in Table~\ref{tbl:mpc}. We compare both styles of regularization. Using our distance as a cost function improves results in all environments.

\input{tables/mpc_ant}
\input{tables/mpc}

\section{Correspondence-wise Averages}

Interestingly, we found that by simply taking averages of context frames under a correspondence-wise loss, as we did in Section~\ref{sect:toy_experiment}, we were able to produce images that resembled frame interpolations. This leads to a frame interpolation method that requires no training at all (although the flow network must be pretrained). We show a duplicate of~\tbl{tbl:frames} in~\tbl{tbl:frames_withavg} with two additional rows: a naive untrained baseline that warps an image by half its flow, and correspondence-wise averages. We see that our image averaging method outperforms the naive baseline, and surprisingly outperforms some supervised methods as well.

\input{tables/frame_interp}

\section{Human Study Details} \label{apdx:amt}

Our perceptual studies are based on the implementation of Zhang~\etal~\cite{zhang2016colorful}, which was also used for evaluating video prediction by Lee~\etal~\cite{lee2018savp}. We used a two-alternative forced choice test. Participants were presented with pairs of videos: one real and one synthesized. The video contained the context (i.e. the ground truth frames given as input) followed by the generated or real frames. The two videos were randomly placed on the left or right side of the screen and shown one after the other, after which two buttons appeared. Participants were then given unlimited time to choose the side the fake video appeared on by clicking one of the two buttons. They were given 5 practice pairs, and then 50 real pairs. The studies were carried out on Amazon Mechanical Turk and we used 50 participants for every ablation or model we evaluated, for a total of 2500 pairs per ablation or model.

\section{Smoothness Trade-offs.}\label{apdx:theory}

In~\sect{sect:toy_experiment} we discuss how our method is not {\em invariant} to position. That is, our method penalizes when objects are in incorrect locations as shown in Figure~\ref{fig:toy_problem}. Analytically, we hypothesize that this is because the underlying flow model makes an explicit trade-off between reconstruction quality and smoothness of the estimated flow.

For example, if we assume a Horn-Schunck-style~\cite{hornschunck1981hs,sun2010secrets} optical flow method, the flow can be written analytically as:
\begin{equation*}
\small
   \mathcal{F}(\im_1, \im_2) = \argmin_\flow ||\operatorname{warp}(\im_1, \flow) - \im_2||_1 + \lambda ||\nabla \flow||_1.
\end{equation*}
Using an $L_1$ base loss, our loss (Eq.~\ref{eq:warpdist}) can be written as:
{\small 
\begin{equation*}
\small
       \mathcal{L}_{C} =  \left\Vert \imgt -    \operatorname{warp}\left(\imest, \left [\argmin_\flow ||\imgt - \operatorname{warp}(\imest, \flow)||_1 + \lambda ||\nabla \flow||_1 \right ]\right) \right\Vert_1
\end{equation*}}
for predicted image $\imest$ and ground truth image $\imgt$.

The optical flow will generally not produce a warp that obtains perfect reconstruction (i.e. one where $\mathcal{L}_{C} = 0$) since it pays a penalty for complex flows, instead choosing a solution that trades off between both terms. If, however, the smoothness term were removed by setting $\lambda = 0$, then the model would be invariant to the position of the scene structure. Objects would ``snap" to the correct locations without penalty, and the resulting image prediction algorithm would be insensitive to the correct positions of objects.

%% file: tables/toy_extra.tex
\begin{figure*}[ht!]
    \centering
    \includegraphics[width=\textwidth]{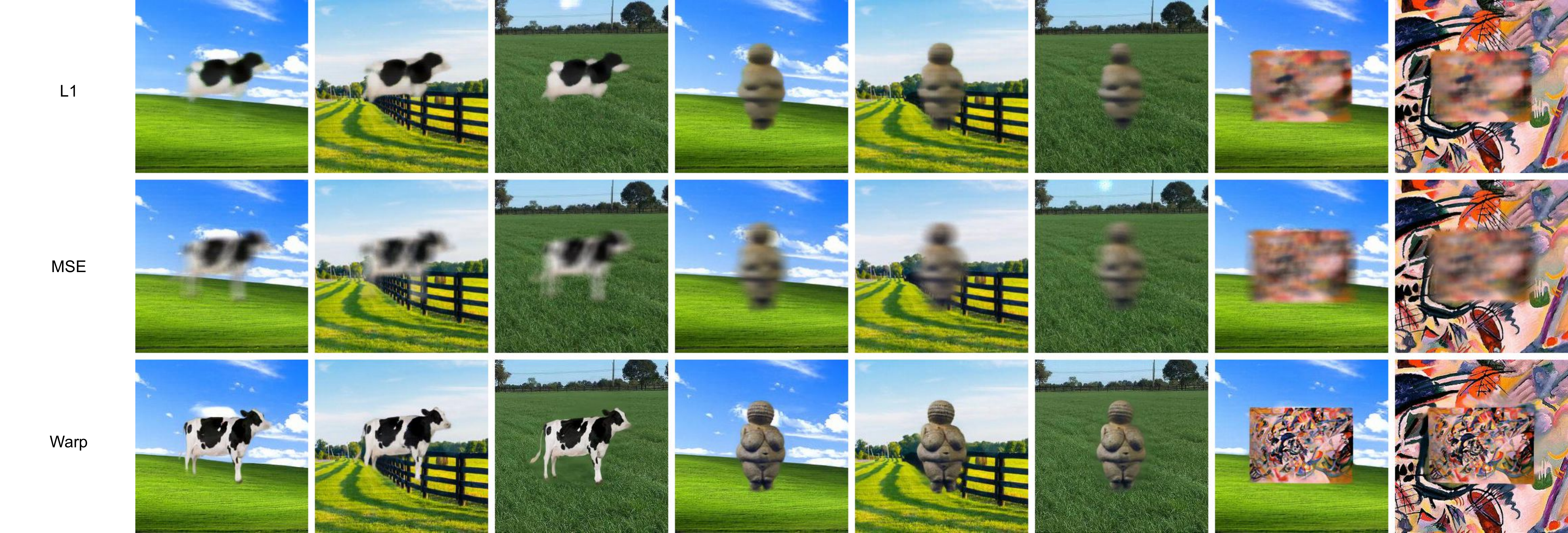}
    \caption{Additional toy experiment results, with various objects and backgrounds. The final row labeled ``warp" is our $ L_1 $ correspondence-wise loss. Our method obtains sharp reconstructions despite the large variation in objects and backgrounds (none of which the flow model was trained for).}
    \label{fig:supp_toy}
\end{figure*}

%% file: tables/architecture.tex
\begin{table}
    \begin{center}
    \resizebox{1.05\linewidth}{!}{
    \begin{tabular}{lcccc}
    \toprule
    Layer & Output Dims & Kernel Size & Padding & Stride \\
    \midrule
    \texttt{input} & $(3, T, H, W) $ & & & \\
    \texttt{conv3D\_1} & $(64, T, H, W) $ & $(3, 7, 7)$ & $(1, 3, 3) $ & $(1, 1, 1) $ \\
    \texttt{conv3D\_2} & $(128, T, H/2, W/2) $ & $(3, 3, 3) $ & $(1, 1, 1) $ & $(1, 2, 2) $ \\
    \texttt{conv3D\_3} & $(256, T, H/4, W/4) $ & $(3, 3, 3) $ & $(1, 1, 1) $ & $(1, 2, 2) $ \\
    \texttt{conv3D\_4} & $(512, T, H/8, W/8) $ & $(3, 3, 3) $ & $(1, 1, 1) $ & $(1, 2, 2) $ \\
    \texttt{res3D} $\times 2$ & $(512, T, H/8, W/8) $ & $(3, 3, 3) $ & $(1, 1, 1) $ & $(1, 1, 1) $ \\
    \texttt{slice\_last} & $(512, H/8, W/8) $ & & & \\
    \texttt{res2D} $\times 4$ & $(512, H/8, W/8) $ & $(3, 3) $ & $(1, 1) $ & $(1, 1) $ \\
    \texttt{nn\_upsample} & $(512, H/4, W/4) $ & & & \\
    \texttt{conv\_1} & $(256, H/4, W/4) $ & $(3, 3) $ & $(1, 1) $ & $(2, 2) $ \\
    \texttt{nn\_upsample} & $(256, H/2, W/2) $ & & & \\
    \texttt{conv\_2} & $(128, H/2, W/2) $ & $(3, 3) $ & $(1, 1) $ & $(2, 2) $ \\
    \texttt{nn\_upsample} & $(64 , H, W) $ & & & \\
    \texttt{conv\_3} & $(64, H, W) $ & $(3, 3) $ & $(1, 1) $ & $(2, 2) $ \\
    \texttt{conv\_4} & $(3, H, W) $ & $(7, 7)$ & $(3, 3) $ & $(1, 1) $ \\
    \bottomrule
    \end{tabular}
    }
    \end{center}
    
    \vspace{-13pt}

\caption{{\bf Architecture}. The architecture used for our simple frame prediction model. \texttt{slice\_last} outputs the last frame of the input sequence.}
\label{tbl:architecture}
\end{table}

%% file: tables/frame_interp_skip.tex
\begin{table}
    \begin{center} \small
    \resizebox{.85\linewidth}{!}{
    \setlength\tabcolsep{1.5pt}
    \begin{tabularx}{\linewidth}{Xcccc}
    \toprule
    Method & Frames & LPIPS ($\downarrow$) & PSNR ($\uparrow$) & SSIM ($\uparrow$) \\
    \midrule
    FLAVR ~\cite{kalluri2021flavr} & 2 & 0.1509 & 25.99 & 0.880\\
    FLAVR + corr. & 2 & 0.1152 & 25.96 &0.880 \\
    \bottomrule
    \end{tabularx}%
    }
    \end{center}
    
    \vspace{-2mm}
    \caption{{\bf Frame interpolation evaluations.} Our additional interpolation experiments, with ``FLAVR + corr" being the FLAVR model with a correspondence-wise $L_1$ loss. These models were trained with a 3x longer time difference between input frames.  \vspace{-4mm}}
    \label{tbl:frames_skip}
\end{table}

%% file: tables/toy_loss.tex
\begin{figure}[h]
    \centering
    \includegraphics[width=.95\linewidth]{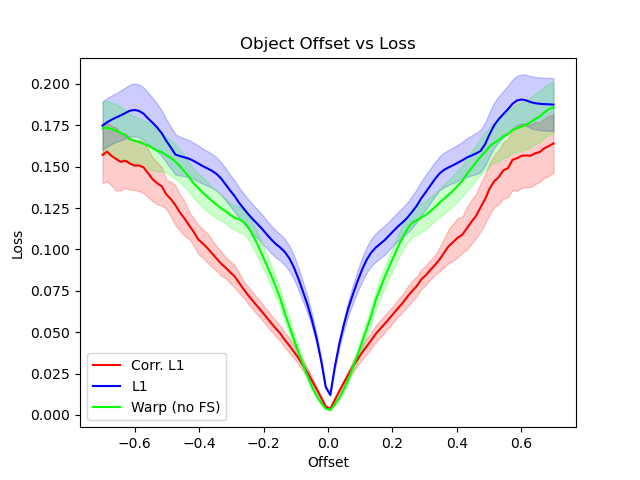}
    \caption{Plots of object offset vs loss. ``Warp" is an $L_1$ correspondence-wise loss, and ``Warp (no FS)" is an $L_1$ correspondence-wise loss without flow scaling. Note that flow scaling makes the loss more proportional to object offset.}
    \label{fig:flow_scaling}
\end{figure}

%% file: tables/mpc_ant.tex
\begin{figure}
    \centering
    \includegraphics[width=\linewidth]{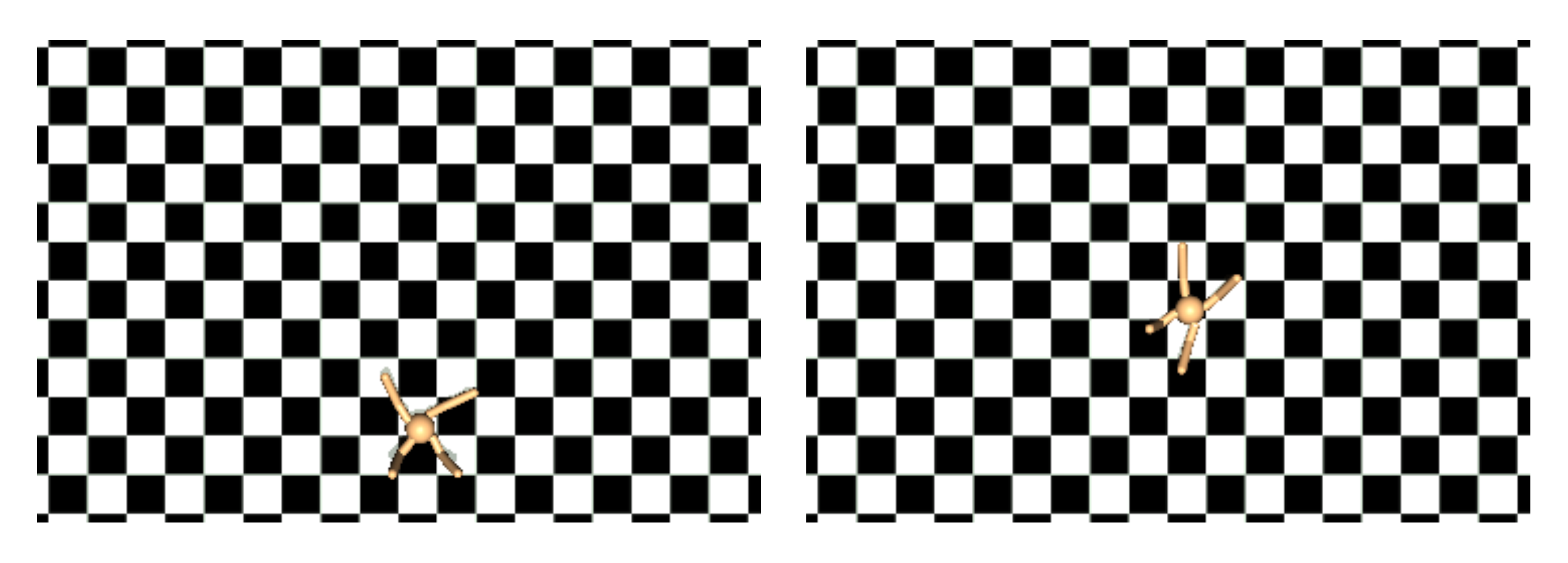}
    \vspace{-2.5mm}
    \caption{{\bf MPC Environments.} Stills from our Ant environment. Note that taking an L1 loss between the two images cannot reliably give the distance.}
    \label{fig:mpc_ant}
\end{figure}

%% file: tables/mpc.tex
\begin{table}
    \begin{center}
    \resizebox{.85\linewidth}{!}{
    \begin{tabular}{lccc}
    \toprule
    & $L_1$ & Corr. $L_1$ + Scaling & Corr. $L_1$ + Reg. \\
    \midrule
    Cartpole & 5.520 & 1.653 & {\bf 1.623} \\
    Ant & 1.735 & 1.710 & {\bf 1.523} \\
    Humanoid & 1.229 & {\bf 1.155} & 1.176 \\
    \bottomrule
    \end{tabular}
    }
    \end{center}
    
    \vspace{-13pt}

\caption{\small {\bf MPC Experiments}. MPC evaluation results using various loss functions. All metrics are minimum distance to goal in state space over a rollout. Every version of our loss outperforms the $L_1$ cost.}
\label{tbl:mpc}
\end{table}

%% file: tables/frame_interp.tex
\begin{table}
    \begin{center} \small
    \setlength\tabcolsep{1.5pt}
    \begin{tabularx}{\linewidth}{Xcccc}
    \toprule
    Method & Frames & LPIPS ($\downarrow$) & PSNR ($\uparrow$) & SSIM ($\uparrow$) \\
    \midrule
    DVF ~\cite{liu2017dvf} & 2 & - & 27.27 & 0.893\\
    SuperSloMo ~\cite{jiang2018super} & 2 & - & 32.90 & 0.957\\
    SepConv ~\cite{niklaus2017video} & 2 & - & 33.60 & 0.944\\
    CAIN ~\cite{choi2020cain} & 2 & - & 33.93 & 0.964\\
    BMBC ~\cite{park2020bmbcbilateral} & 2 & - & 34.76 & 0.965\\
    AdaCoF ~\cite{lee2020adacof} & 2 & - & 35.40 & 0.971\\
    FLAVR ~\cite{kalluri2021flavr} & 4 & 0.0248 & 36.30 & 0.975\\
    \midrule
    FLAVR ~\cite{kalluri2021flavr} & 2 & 0.0297 & 34.96 & 0.970\\
    FLAVR + corr. & 2 & 0.0268 & 35.13 & 0.970 \\
    \midrule
    Half Flow & 2 & 0.0481 & 29.57 & 0.945 \\
    Averages & 2 & 0.0342 & 30.81 & 0.947 \\
    \bottomrule
    \end{tabularx}%
    \end{center}
    
    \vspace{-2mm}
    \caption{{\bf Frame interpolation evaluations.} A copy of Table~\ref{tbl:frames}, with two additional rows: a half-flow warping baseline, and correspondence-wise averages. \vspace{-4mm}}
    \label{tbl:frames_withavg}
\end{table}